\pdfoutput=1
\documentclass[letterpaper]{article} % DO NOT CHANGE THIS
\usepackage{aaai24}  % DO NOT CHANGE THIS
\usepackage{times}  % DO NOT CHANGE THIS
\usepackage{helvet}  % DO NOT CHANGE THIS
\usepackage{courier}  % DO NOT CHANGE THIS
\usepackage[hyphens]{url}  % DO NOT CHANGE THIS
\usepackage{graphicx} % DO NOT CHANGE THIS
\usepackage{natbib}  % DO NOT CHANGE THIS AND DO NOT ADD ANY OPTIONS TO IT
\usepackage{caption} % DO NOT CHANGE THIS AND DO NOT ADD ANY OPTIONS TO IT
\usepackage{algorithm}
\usepackage{algorithmic}

\usepackage[utf8]{inputenc} % allow utf-8 input
\usepackage[T1]{fontenc}    % use 8-bit T1 fonts
\usepackage{booktabs}       % professional-quality tables
\usepackage{amsfonts}       % blackboard math symbols
\usepackage{nicefrac}       % compact symbols for 1/2, etc.
\usepackage{microtype}      % microtypography
\usepackage[dvipsnames]{xcolor}         % colors
\usepackage{enumitem}
\usepackage{multirow}
\usepackage{amsmath}
\usepackage{mdframed}
\usepackage{subcaption}

\usepackage{newfloat}
\usepackage{listings}
\DeclareCaptionStyle{ruled}{labelfont=normalfont,labelsep=colon,strut=off} % DO NOT CHANGE THIS
\floatstyle{ruled}
\newfloat{listing}{tb}{lst}{}
\floatname{listing}{Listing}
\title{MotionGPT: Finetuned LLMs Are General-Purpose Motion Generators}
\author{
    Yaqi Zhang\textsuperscript{\rm 1,2}, Di Huang\textsuperscript{\rm 3}\thanks{Project leader.}, Bin Liu\textsuperscript{\rm 1,2}\thanks{Corresponding author.}, Shixiang Tang\textsuperscript{\rm 3}, Yan Lu\textsuperscript{\rm 3},\\
    Lu Chen\textsuperscript{\rm 4}, Lei Bai\textsuperscript{\rm 4}, Qi Chu\textsuperscript{\rm 1,2}, Nenghai Yu\textsuperscript{\rm 1,2}, Wanli Ouyang\textsuperscript{\rm 4}
}
\affiliations{
    \textsuperscript{\rm 1}School of Cyber Science and Technology, University of Science and Technology of China\\
    \textsuperscript{\rm 2}CAS Key Laboratory of Electromagnetic Space Information\\
    \textsuperscript{\rm 3}The University of Sydney\\ \textsuperscript{\rm 4}Shanghai AI Laboratory\\
    zhangyq99@mail.ustc.edu.cn, flowice@ustc.edu.cn
}

\newcommand{\gray}[1]{\textcolor[rgb]{0.6,0.6,0.6}{#1}}

\begin{document}

\maketitle
\begin{abstract}
Generating realistic human motion from given action descriptions has experienced significant advancements because of the emerging requirement of digital humans.
While recent works have achieved impressive results in generating motion directly from textual action descriptions, 
they often support only a single modality of the control signal, which limits their application in the real digital human industry.
This paper presents a \textbf{Motion} \textbf{G}eneral-\textbf{P}urpose genera\textbf{T}or (MotionGPT) that can use multimodal control signals, \emph{e.g.,} text and single-frame poses, for generating consecutive human motions by treating multimodal signals as special input tokens in large language models (LLMs).
Specifically, we first quantize multimodal control signals into discrete codes and then formulate them in a unified prompt instruction to ask the LLMs to generate the motion answer.
Our MotionGPT demonstrates a unified human motion generation model with multimodal control signals by tuning a mere 0.4\% of LLM parameters. To the best of our knowledge, MotionGPT is the first method to generate human motion by multimodal control signals, which we hope can shed light on this new direction. Visit our webpage at \url{https://qiqiapink.github.io/MotionGPT/}.

\end{abstract}
\section{Introduction}
Human motion is pivotal in various applications such as video gaming, filmmaking, and virtual reality.
Recent advancements in AI~\cite{saharia2022photorealistic,yu2022scaling,ramesh2022hierarchical,rombach2022high,ramesh2021zero,ouyang2022training,lu2023seeing} have paved the way for novel approaches to motion creation, enabling various control conditions including textual descriptions, music pieces, and human poses. 
However, one significant shortcoming of existing works~\cite{petrovich2022temos,zhang2022motiondiffuse,tevet2022human,petrovich2021action,zhuang2022music2dance} is that they only target a single type of control condition, greatly limiting their applications in the real world, \emph{e.g.,} unable to generate motion sequences conditioned on text descriptions and several keyframe human poses.
To facilitate such applications, it is important to develop a unified human motion generation framework that can efficiently utilize multiple control signals simultaneously.
 
This paper proposes a novel and more unified framework for text-motion generation. 
The framework facilitates the generation of human motions using multiple control conditions, formulated as $output\_motion=f(text, task, input\_motion)$.
Newly added inputs $task$ and $input\_motion$ represent the task and given motion prompts, respectively. 
Here, $task$ indicates the specific task the model should adapt to, while $input\_motion$ provides the keyframe poses corresponding to the given task.
This framework is a departure from traditional text-motion generation models as the introduction of $input\_motion$ enables more precise control.
For example, given an $input\_motion$ and set the $task$ as "generate motion given initial poses", the model should compensate for the subsequent frames of the given frames. 
Such a framework offers a more practical and comprehensive solution for human motion generation, where task instructions and multimodal conditions can flexibly control motion generation.

The challenge of building a model to complete such (text, motion)-motion generation task lies in understanding multimodal control conditions and generating human motions with varying motion lengths and richer patterns. We argue that these challenges can be naturally resolved by adapting from LLMs for the following reasons. First, recent studies have demonstrated that LLMs can understand multimodal inputs, \emph{e.g.,} images~\cite{zhu2023minigpt,du2023vision,li2023blip, liu2023llava, ye2023mplugowl} and videos~\cite{2023videochat}, through a lightweight adapter~\cite{hu2021lora}. 
Therefore, we expect the LLMs can also understand motion sequences with an appropriate adapter. Second, LLMs can provide diverse human motion contexts for motion generation because they have encoded diverse motion patterns from extensive large-scale text data. This enables our motion generator fine-tuned from LLMs can produce motions with rich patterns. Third, since LLMs output tokens aggressively, producing human motion with flexible sequences is no longer an obstacle.

To this end, we propose a \textbf{Motion} \textbf{G}eneral-\textbf{P}urpose genera\textbf{T}or (MotionGPT) by fine-tuning an LLM following designed instructions. Specifically, MotionGPT first maps human poses into discrete motion codes via the pre-trained motion VQ-VAE and then generates instructions by combining codes from language prompts and motion prompts. The LLMs are fine-tuned by answering the correct human pose sequences to the instructions in an efficient way of well-known LoRA adaptation. The designed motion instruction tuning framework can incorporate pose sequence information into the fine-tuned large language model while taking advantage of strong motion priors in the original large language model.

We conduct extensive experiments on the HumanML3D~\cite{guo2022generating} and KIT-ML~\cite{plappert2016kit} datasets,  demonstrating MotionGPT has a strong ability for motion generation with multiple control conditions.
Remarkably, MotionGPT achieves this with a significantly small set of training parameters (33 M), and in less training time (about 4 hours, or just 10\% of the time taken by other methods). 
We observe that joint training under multiple control instructions outperforms training with a single type of control signal, showing the effectiveness of our unified motion generation training paradigm.
Our contributions can be summarized as follows:

\begin{figure*}[t]
    \centering
    \includegraphics[width=\textwidth,trim=30 20 30 30,clip]{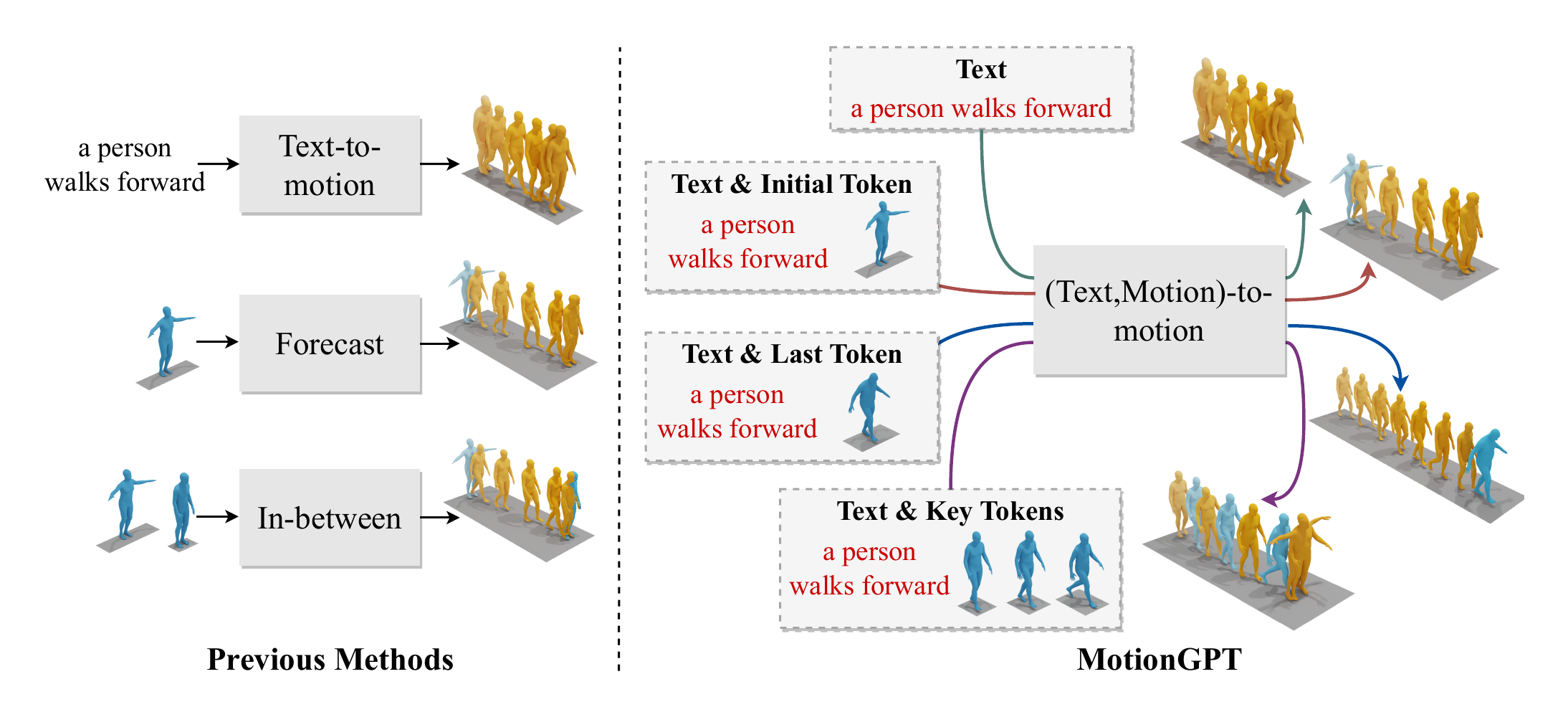}
    \caption{
    This work proposes a novel human motion generation method via fine-tuned LLMs, named MotionGPT. Compared with previous methods, MotionGPT has the unique ability to accept multiple control conditions and solve various motion generation tasks using a unified model.
    }
    \label{fig:teaser}
    \vspace{-0.2cm}
\end{figure*}

\begin{itemize}[leftmargin=*]
\item We introduce a novel model, MotionGPT, for generating human motions, which allows for multiple types of control during the generation process. 
To the best of our knowledge, MotionGPT is the first method for using both text and poses as conditions. It supports generating subsequent, preceding, or `in-betweening' motions using a single and unified model.
\item We demonstrate that a pre-trained LLM can be readily tuned to function as a human motion generator, suggesting the potential for directly utilizing LLMs for human motion generation.
\item We present a comprehensive set of experiments, showcasing the effectiveness of our proposed MotionGPT with multiple types of control signals. Experimental results also indicate that using a more powerful LLM results in superior motion generation quality, indicating that further advancements in LLM technology could substantially enhance the performance of MotionGPT in the future.
\end{itemize}
\section{Related Work}

\paragraph{Large language models}
Recently, large language models~\cite{devlin2018bert,radford2018improving,radford2019language,brown2020language,OpenAI2023GPT4TR,touvron2023llama} have been developed dramatically, \emph{e.g.,} BERT~\cite{devlin2018bert}, GPT~\cite{radford2018improving}, and Google T5~\cite{raffel2020exploring}.
These models, such as GPT-4~\cite{OpenAI2023GPT4TR}, demonstrate exceptional performance on various linguistic tasks, thanks to the extensive training data (45 gigabytes in the case of GPT-4) and the large number of parameters they leverage.
Previously, language models were task-specific, focusing on areas such as translation and sentiment analysis. However, recent developments, like ChatGPT, have expanded the capability of these models. Based on GPT-4, ChatGPT can interact with humans, showcasing its strong natural language understanding abilities. This effectiveness has opened up possibilities for a myriad of downstream tasks achieved through fine-tuning these LLMs.
However, fine-tuning such models, considering their extensive parameters, is a challenging task. To address this issue, efficient fine-tuning strategies have been proposed, including prompt tuning~\cite{lester2021power,liu2021p,hu2021knowledgeable}, adapters~\cite{houlsby2019parameter,he2021effectiveness,le2021lightweight}, and LoRA~\cite{hu2021lora}.
Our work draws inspiration from the recent progress in LLMs, but it also addresses a distinct problem by introducing a new modality into the LLMs.

\paragraph{Human motion generation}
Motion generation~\cite{tevet2022motionclip,habibie2017recurrent,petrovich2021action,li2017auto,zhang2022motiondiffuse,guo2020action2motion,tevet2022human,petrovich2022temos,li2021ai} is a long-history task that can be conditioned on various conditions, such as motion description, actions, and music.
For instance, HP-GAN~\cite{barsoum2018hp} and \cite{martinez2017human} utilize a sequence-to-sequence model to anticipate future poses based on prior poses. 
ACTOR~\cite{petrovich2021action} employs a transformer VAE for both unconditional and action-based generation.
TRAJEVAE~\cite{kania2021trajevae}, when supplied with an initial pose and a trajectory, can generate a motion sequence that follows the given path.
In recent years, text-conditional motion generation has garnered significant attention. This approach focuses on generating human motion sequences conditioned on textual descriptions. 
TEMOS~\cite{petrovich2022temos} proposes a VAE model that learns a shared latent space for both motion and text. 
MotionDiffuse~\cite{zhang2022motiondiffuse} integrates a diffusion model into the text-to-motion generation framework and accomplishes impressive results. 
MDM~\cite{tevet2022human}, aiming to enhance motion-text consistency, uses CLIP~\cite{radford2021learning} as the text encoder to incorporate more robust text priors into the model.
In comparison to previous methods, our work, MotionGPT, stands out as the first unified motion generation model that supports multimodal controls.
\section{MotionGPT: A Motion General-Purpose Generator}
MotionGPT proposes a \textbf{Motion} \textbf{G}eneral-\textbf{P}urpose genera\textbf{T}or controlled by multimodal conditions, \emph{i.e.,} texts and human poses in keyframes. Our motivation is to formulate human motion as a problem of asking the Large Language Model to generate desirable human motions according to task prompts and control conditions. 
Specifically, we quantize motion controls into discrete codes using the widely-used VQ-VAE~\cite{van2017neural}. 
Motion discrete codes, text control conditions, and designed task instructions are then organized into a unified question template for the LoRA-finetuned LLM to generate a human motion sequence answer. 
Following the typical framework of instruction tuning, we leverage cross-entropy loss to supervise the LoRA adapter. More importantly, our MotionGPT can address not only existing human motion generation tasks, \emph{e.g.,} text-to-motion generation, but also new motion generation tasks by simply adjusting task instructions, showing the potential of MotionGPT as a generic baseline framework for motion generation.

\subsection{Motion Code Generation}
\label{sec: motion code generation}
VQ-VAE proposed in \cite{van2017neural} enables the model to learn discrete representations for generative models. Given a human pose $\mathbf{m}$, the motion VQ-VAE can be trained by the reconstruction loss, the embedding loss and the commitment loss, \emph{i.e.,}
\begin{equation}
\begin{aligned}
    \mathcal{L}_{\text{VQVAE}} = ||\mathcal{D}(\mathcal{E}(\mathbf{m}))-\mathbf{m}||^2 + \|\text{sg}[\mathcal{E}(\mathbf{m})] - \mathbf{e}\|^2_2 \\
    + \beta\|\mathcal{E}(\mathbf{m}) - \text{sg}[\mathbf{e}]\|^2_2,
\end{aligned}
\end{equation}
where $\mathcal{E}$, $\mathcal{D}$ are the motion encoder and the motion decoder, respectively. $\text{sg}$ indicates the stop gradient operation. Here, the estimated embedding $\mathbf{e}$ after qunatization can be found by searching the nearest embedding in a learnable codebook $\mathcal B=\{b_1, b_2, ..., b_N\}$, where $N$ is the size of the codebook, which can be mathematically formulated as
\begin{equation}
    \mathbf{e} = \mathop{\arg\min}\limits_{b_k\in\mathcal B}\|\mathcal{E}(\mathbf{m})-b_k\|_2.
\end{equation}
Based on the estimation latent representation $\mathbf{e}$ of the motion $\mathbf{m}$, the reconstructed human pose $\mathbf{\hat{m}}$ can be produced by the decoder of VQ-VAE and the motion code $p$ of human pose $\mathbf{m}$ can be calculated as the index of its nearest embedding in the codebook, \emph{i.e.,} 
\begin{equation}
     \mathbf{\hat{m}} = \mathcal{D}(\mathbf{e}), \quad p = \mathop{\arg\min}\limits_{k}\|\mathcal{E}(\mathbf{m})-b_k\|_2.
    \label{eq: vq-vae}
\end{equation}

\subsection{Instruction Generation}
\label{sec: instruction generation}
In MotionGPT, we design instructions that combine task prompts and control conditions to enable (text, motion)-motion generation tasks. Specifically, given the task prompts $\mathcal{T}=\{t_1, t_2, ..., t_{n_t}\}$, the text control conditions $\mathcal{X}=\{x_1, x_2, ..., x_{n_x}\}$ and the pose control conditions $\mathcal{P}=\{p_1, p_2, ..., p_{n_p}\}$ where $n_t$, $n_x$ and $n_p$ are the number of codes in $\mathcal{T}$, $\mathcal{X}$ and $\mathcal{P}$, the instruction $\mathcal{I}$ is formulated as
\begin{mdframed}[backgroundcolor=gray!20]
\gray{\% General control conditions format} \\
Control Conditions: \{Text control conditions $\mathcal{X}$ $<$$x_1, x_2, ..., x_{n_x}$$>$\} \{Pose control conditions $\mathcal{P}$ $<$$p_1, p_2, ..., p_{n_p}$$>$\} \\
\gray{\% General instruction format } \\
Instruction $\mathcal{I}$:
\{Task Prompts $\mathcal{T}$ $<$$t_1, t_2, ..., t_{n_t}$$>$\} 
\{Control Conditions\}
\end{mdframed}
Here, the pose control conditions $\mathcal{P}=\{p_1, p_2, ..., p_{n_p}\}$ presents pose codes, generated by using the same motion VQ-VAE mentioned earlier.
Consequently, the entire instruction $\mathcal{I}$ can be regarded as a sequence of specialized text inputs.
By generating different motion instructions, our MotionGPT can address existing human motion generation tasks and new human motion generations.

\begin{figure*}[t]
    \centering
    \includegraphics[width=\textwidth,trim=0 70 60 10,clip]{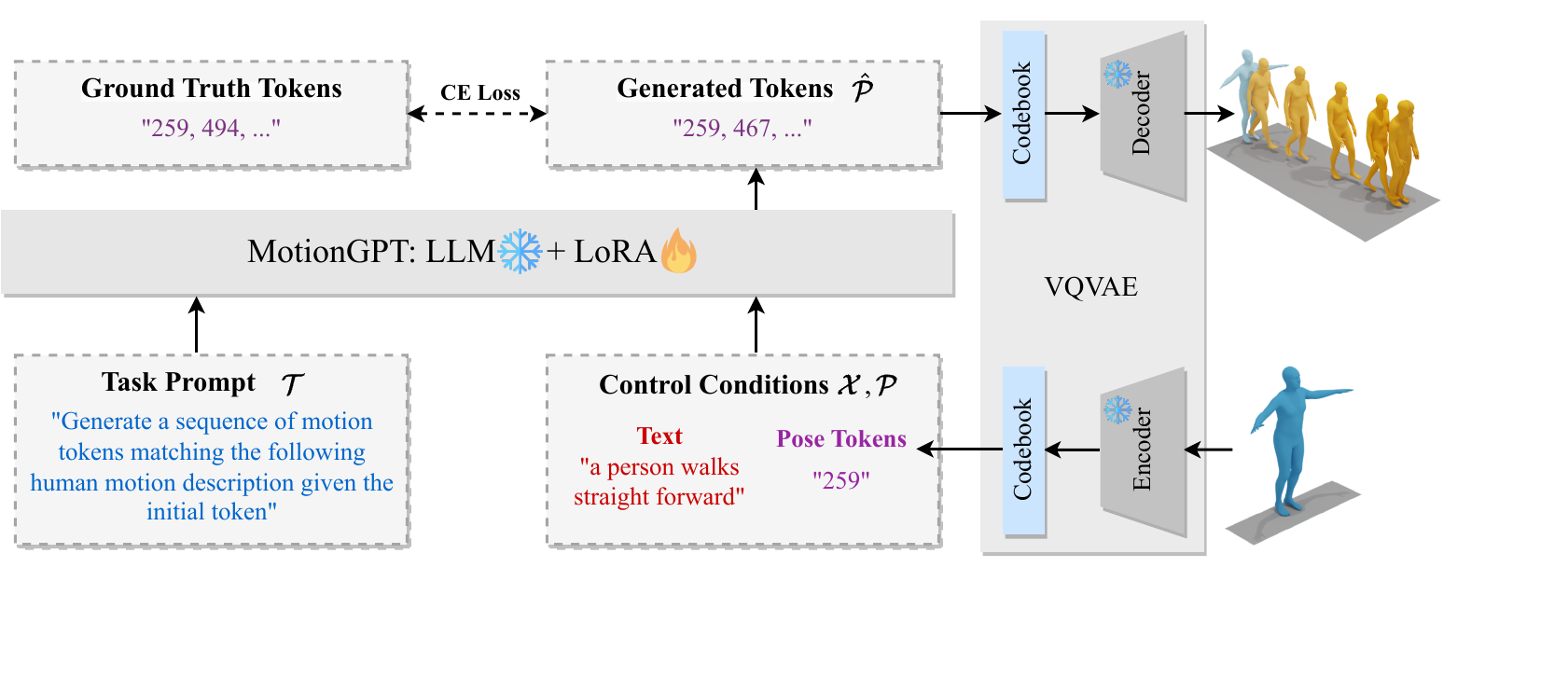}
    \caption{
    The pipeline of MotionGPT, a Motion General-Purpose generaTor.
    Given text and poses as an input example, we organize task descriptions (Instruction) and multiple control conditions (Input) within a question template. 
    MotionGPT fine-tunes an LLM to generate the corresponding motion answer, which can then be decoded into human motions using a VQ-VAE decoder.
    }
    \label{fig:pipeline}
\end{figure*}

\subsection{Fine-tuning LLM by Motion Instructions}
\label{sec: Finetuning LLM by Motion Instructions}
Instruction tuning~\cite{wei2021finetuned} enables LLMs to handle various generation tasks by asking the LLM questions in different instructions. Therefore, we design various instructions that combine both task descriptions and control conditions to fine-tune large language model by the widely-used and efficient Low-Rank Adaptation (LoRA)~\cite{hu2021lora}. Specifically, given a large language model $\mathcal{F}$, the general template of our instructions $\mathcal{I}$ and the answer of the LLM $\mathcal{\hat{P}}=\mathcal{F}(\mathcal{I})$ are formulated as 

\begin{mdframed}[backgroundcolor=gray!20]
Below is an instruction that describes a task, paired with an input that provides further context. Write a response that appropriately completes the request. \\
\gray{\% Task Prompts: Code sequences of Task Prompts} \\
\gray{\% Control Conditions: Code sequences of Control Conditions} \\
\textbf{Instruction $\mathcal{I}$:} \{Task Prompts $\mathcal{T}$\} \{Control Conditions\}\\
\textbf{Answer $\mathcal{\hat{P}}$: } \{Sequences of Human Motions\}
\end{mdframed}

The answer of LLM $\mathcal{\hat{P}}= \{\hat{p}_1, \hat{p}_2, ..., \hat{p}_{n_{\hat{p}}}\}$ is a series of generated motion codes, which can be decoded to human motion using Eq.~\ref{eq: vq-vae}.

Similar to most language models, we employ cross-entropy loss which constrains the similarity between estimated and ground-truth tokens, to fine-tune LLMs by LoRA, which can be presented as 
\begin{equation}
\mathcal{L}_{lora} = {\rm CE}(\hat{\mathcal{P}}, \hat{\mathcal{P}^{gt}}),
\end{equation}
where $\hat{\mathcal{P}^{gt}}$ is the motion codes of ground-truth motions calculated by Eq.~\ref{eq: vq-vae} and $\hat{\mathcal{P}}$ is the motion codes predicted by the LLM $\mathcal{F}$.

\subsection{Generalization to Existing and New Tasks}
\label{sec: generalization to existing and new tasks}
Leveraging the general template given before, our MotionGPT is capable of being a general-purpose motion generator, supporting various generation tasks.
Specifically, for existing text-to-motion generation setting, MotionGPT address it by constructing following instruction $\mathcal{I}$:
\begin{mdframed}[backgroundcolor=gray!20]
\textbf{Instruction ($\mathcal{I}):$} \{\gray{Task Prompts:} "Generate a sequence of motion tokens matching the following human motion description."\} \{\gray{Control Conditions:} Text control condition $\mathcal{X}$\}
\end{mdframed}

By adjusting instructions, MotionGPT can be easily adapted to multiple control conditions, \emph{e.g.} text and an arbitrary number of human poses:

\begin{mdframed}[backgroundcolor=gray!20]
\textbf{Instruction ($\mathcal{I}):$} \{\gray{Task Prompts:} "Generate a sequence of motion tokens matching the following human motion description given the init/last/key pose tokens."\} \{\gray{Control Conditions:} Text control condition $\mathcal{X}$ $<$Motion Token$>$ Pose control conditions $\mathcal{P}$ $<$/Motion Token$>$\}
\end{mdframed}
\section{Experiment}

\subsection{Datasets and Evaluation Metrics}\label{sec:dataset}
\paragraph{Datasets}
We apply two widely-used datasets, HumanML3D~\cite{guo2022generating} and KIT-ML~\cite{plappert2016kit} for evaluation.

\begin{table*}[ht]
    \centering
    \begin{tabular}{l|ccc|ccc}
        \toprule
             \multirow{2}{*}{Methods} & \multicolumn{3}{c|}{HumanML3D} & \multicolumn{3}{c}{KIT-ML} \\
            & {FID $\downarrow$} & {MM Dist $\downarrow$} & {Diversity $\uparrow$} & {FID $\downarrow$} & {MM Dist $\downarrow$} & {Diversity $\uparrow$} \\
        \midrule
            Real motion & 0.002 & 2.974 & 9.503 & 0.031 & 2.788 & 11.08 \\
        \midrule
            TEMOS~\cite{petrovich2022temos} & 3.734 & 3.703 & 8.973 & 3.717 & 3.417 & 10.84 \\
            TM2T~\cite{guo2022tm2t} & 1.501 & 3.467 & 8.589 & 1.501 & 3.467 & 8.589 \\
            T2M~\cite{guo2022generating} & 1.087 & 3.347 & 9.175 & 3.022 & 3.488 & 10.72 \\
            MotionDiffuse~\cite{zhang2022motiondiffuse} & 0.630 & {\bf 3.113} & 9.410 & 1.954 & {\bf 2.958} & {\bf 11.10} \\
            MDM~\cite{tevet2022human} & 0.544 & 5.566 & 9.559 & \underline{0.497} & 9.191 & 10.85 \\
            MLD~\cite{chen2023mld} & \underline{0.473} & 3.196 & \underline{9.724} & \bf{0.404} & 3.204 & 10.80 \\
            T2M-GPT~\cite{zhang2023t2m} & \bf{0.116} & \underline{3.118} & \bf{9.761} & 0.514 & \underline{3.007} & \underline{10.92} \\
        \midrule
            MotionGPT-13B (Ours) & 0.567 & 3.775 & 9.006 & 0.597 & 3.394 & 10.54 \\
        \bottomrule
    \end{tabular}
    \caption{Comparisons of text-to-motion generation with the state-of-the-art methods on HumanML3D and KIT-ML test set. MotionGPT-13B achieves comparable performance on all metrics. Bold and underline indicate the best and the second best result.}
    \label{tab:humanml3d}
\end{table*}

\begin{figure*}[h!]
    \centering
    \includegraphics[width=.95\textwidth,trim=210 180 130 0,clip]{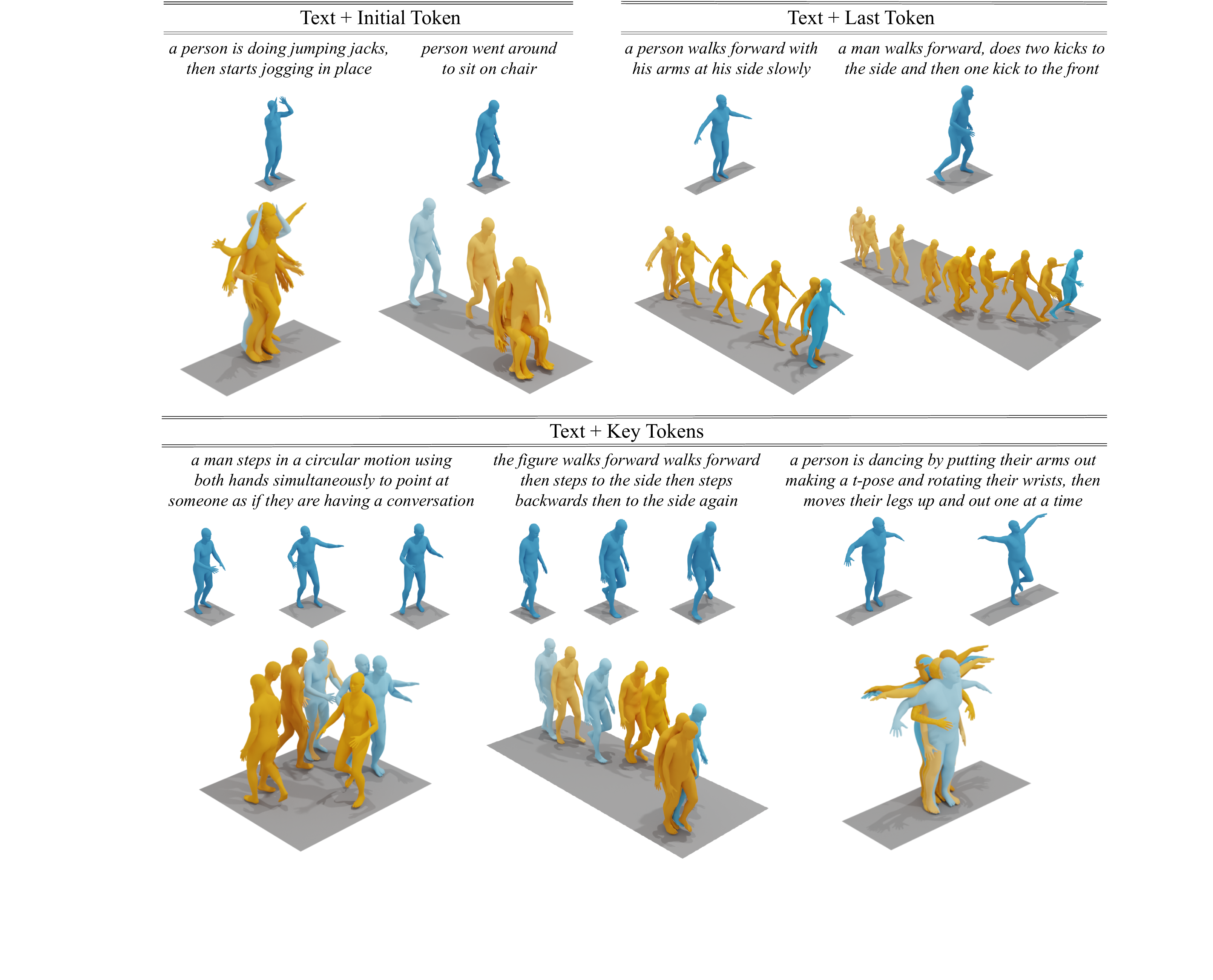}
    \caption{
    Generated motion by MotionGPT with multiple control conditions on HumanML3D. 
    }
    \label{fig:(Text+Pose)-to-motion}
    \vspace{-0.2cm}
\end{figure*}

\paragraph{Evaluation metrics}
Our evaluation comprises two categories of metrics.
Firstly, to assess the quality of the generated motion, we adopt evaluation metrics consistent with previous methods. These include the \textit{Frechet Inception Distance (FID)}, \textit{Multi-modal Distance (MM Dist)}, \textit{R-Precision} (calculating the Top-1/2/3 motion-to-text retrieval accuracy), and the \textit{Diversity} metric. These metrics collectively provide a robust indication of both the realism and diversity of the generated motion.

Secondly, we introduce new metrics tailored to our proposed motion generation setting, including \textit{Reconstruction Loss (Recon)} and \textit{Velocity Loss (Vel)}. Specifically, these metrics aim to measure the consistency between the provided pose conditions and the generated motion.

More information about datasets, proposed new metrics, and implementation details are included in the supplementary material~\cite{zhang2023motiongpt}.

\begin{table}[t]
    \centering
    \small
    \begin{tabular}{lccc}
        \toprule
             Methods & {FID $\downarrow$} & {MM Dist $\downarrow$} & {Diversity $\uparrow$} \\
        \midrule
            \multicolumn{4}{c}{HumanML3D} \\
        \midrule
            Text-only & 0.567 & 3.775 & 9.006 \\
            Text + Initial poses & 0.520 & 3.844 & 9.588 \\
            Text + Last poses & 0.591 & 3.718 & 9.251 \\
            Text + Random poses & 0.367 & 3.598 & 9.176 \\
        \midrule
            \multicolumn{4}{c}{KIT-ML} \\
        \midrule
            Text-only & 0.597 & 3.394 & 10.54 \\
            Text + Initial poses & 0.664 & 3.445 & 10.39 \\
            Text + Last poses & 0.856 & 3.336 & 10.58 \\
            Text + Random poses & 0.671 & 3.411 & 10.76 \\
        \bottomrule
    \end{tabular}
    \caption{Motion generation quality on HumanML3D and KIT-ML test set for diverse control conditions.}
    \label{tab:motion-quality}
    % \vspace{-0.2cm}
\end{table}

\subsection{Comparisons for Motion Generation with Multiple Control Conditions}

In this section, we conduct four different generation experiments with 1) text as the condition, 2) text and initial pose as the condition, 3) text and last pose as the condition, and 4) text and random keyframe pose as the condition. 
For both 2) and 3), we use 4 frame poses as the input pose condition; While for 4), we random sample 12 to 20 frame poses as the pose condition.

\begin{figure*}[ht]
    \centering
    \includegraphics[width=\textwidth,trim=25 110 60 0,clip]{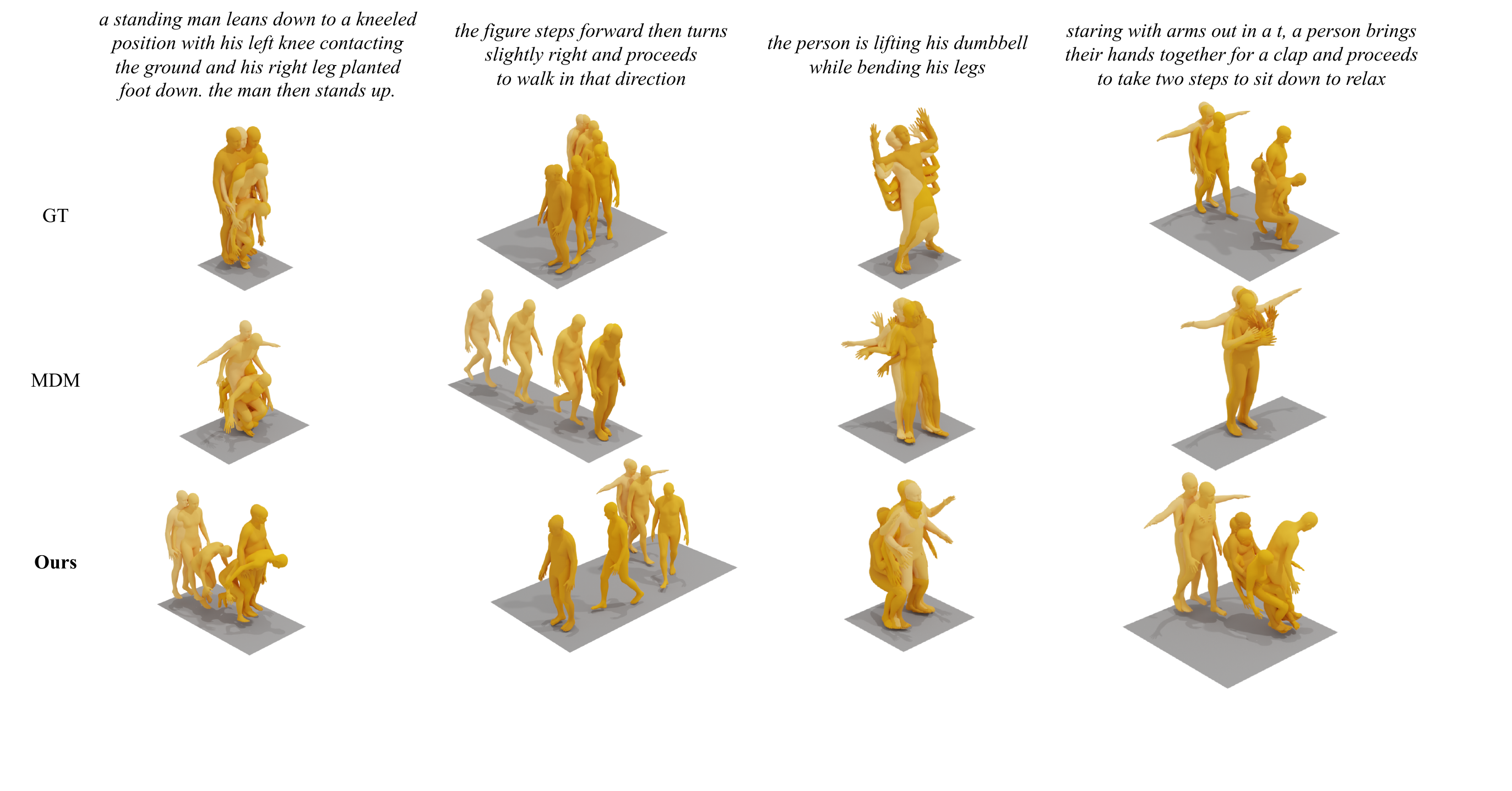}
    \caption{
    Qualitative comparison of the state-of-the-art motion generation method MDM with text-only conditions on HumanML3D. 
    }
    \label{fig:Text-to-motion}
\end{figure*}

The quantitative results of motion quality are depicted in Tab.~\ref{tab:humanml3d} and Tab.~\ref{tab:motion-quality}.
As illustrated in Tab.~\ref{tab:humanml3d}, our proposed model, MotionGPT, exhibits a performance that is competitive with state-of-the-art methods for text-to-motion generation.
Specifically, MotionGPT consistently achieves comparable results across all metrics on both HumanML3D~\cite{guo2022generating} and KIT-ML~\cite{plappert2016kit} datasets.
% As illustrated in Tab.~\ref{tab:humanml3d}, our proposed model, MotionGPT, consistently achieves comparable results with state-of-the-art methods for text-to-motion generation across all metrics on both HumanML3D~\cite{guo2022generating} and KIT-ML~\cite{plappert2016kit} datasets.
In addition to text conditions, MotionGPT can also incorporate human poses as a secondary control modality and the motion quality results are demonstrated in Tab.~\ref{tab:motion-quality}. The adoption of additional control conditions, such as initial, last, or key tokens, does not compromise the quality of the generated motions. In some instances, such as when provided with initial or key tokens, MotionGPT even outperforms its text-only counterpart from $0.567$ to $0.520$ or $0.367$ under FID metric on HumanML3D, demonstrating its robustness and flexibility in handling diverse control modalities.
Nevertheless, a slight decrease in performance is observed when the model is given the final pose as input, which is in line with our expectations, as generating motions with a predetermined end pose presents an inherently greater challenge. Despite this, MotionGPT's performance remains commendable, further affirming its capability to generate high-quality, diverse motions under various control conditions.

We present visualization results in Fig.~\ref{fig:(Text+Pose)-to-motion} and Fig.~\ref{fig:Text-to-motion}. 
As the Fig.~\ref{fig:(Text+Pose)-to-motion} shown, the motions generated by our model exhibit a notable alignment with the provided poses, while also displaying a consistent adherence to the textual descriptions.
For the text-to-motion generation task, we compare our model, MotionGPT, with the MDM, as depicted in Fig.~\ref{fig:Text-to-motion}. Our model demonstrates superior text-consistency and text-completeness compared to MDM~\cite{tevet2022human}. The motions generated by the MDM model often tend to align with only the initial segment of the description, ignoring the latter half. In contrast, our approach exhibits a more comprehensive understanding of the motion descriptions by leveraging the powerful capabilities of LLMs, thus generating more complete and nuanced motion sequences.

\begin{table*}[t]
    \centering
    \small
    \begin{tabular}{l|cccccc}
         \toprule
             \multirow{2}{*}{Pre-trained Model} & \multirow{2}{*}{FID $\downarrow$} & \multirow{2}{*}{MM Dist $\downarrow$} & \multicolumn{3}{c}{R-Precision $\uparrow$} & \multirow{2}{*}{Diversity $\uparrow$} \\
            & & & Top-1 & Top-2 & Top-3 & \\
        \midrule
            LLaMA w/o pre-trained & 26.01 & 8.445 & 0.032 & 0.067 & 0.106 & \bf{9.745} \\
            LLaMA-7B & 0.590 & 3.796 & 0.376 & 0.553 & 0.657 & 9.048 \\
            LLaMA-13B & \bf{0.542} & \bf{3.584} & \bf{0.411} & \bf{0.594} & \bf{0.696} & 9.311 \\
        \bottomrule
    \end{tabular}
    \caption{Evaluation of text-to-motion generation using different pre-trained LLaMA on HumanML3D validation set. Bold indicates the best result.
    }
    \label{tab:pretrained-llm}
\end{table*}

\begin{table*}[!h]
    \centering
    \small
    \begin{tabular}{l|c|cccccc}
         \toprule
             \multirow{2}{*}{Task} & Training & \multirow{2}{*}{FID $\downarrow$} & \multirow{2}{*}{MM Dist $\downarrow$} & \multicolumn{3}{c}{R-Precision $\uparrow$} & \multirow{2}{*}{Diversity $\uparrow$} \\
            & Strategy & & & Top-1 & Top-2 & Top-3 & \\
        \midrule
            Text & \multirow{4}{*}{Separate} & 0.670 & 4.267 & 0.299 & 0.469 & 0.577 & 9.745 \\
            \quad + Initial token & & 0.756 & 3.802 & 0.374 & 0.556 & 0.658 & 9.148 \\
            \quad + Last token & & 1.409 & 4.516 & 0.290 & 0.446 & 0.564 & 8.771 \\
            \quad + Key tokens & & 0.702 & 3.690 & 0.370 & 0.546 & 0.668 & 8.974 \\
        \bottomrule
        \toprule
            Text & \multirow{4}{*}{Joint} & $0.590^{-.180}$ & $3.796^{-.471}$ & $0.376^{+.077}$ & $0.553^{+.084}$ & $0.657^{+.080}$ & $9.048^{-.697}$ \\
            \quad + Initial token & & $0.493^{-.263}$ & $3.750^{-.052}$ & $0.384^{+.010}$ & $0.564^{+.008}$ & $0.666^{+.008}$ & $9.378^{+.230}$ \\
            \quad + Last token & & $0.646^{-.763}$ & $3.675^{-.841}$ & $0.393^{+.103}$ & $0.577^{+.131}$ & $0.681^{+.117}$ & $9.030^{+.259}$ \\
            \quad + Key tokens & & $0.390^{-.663}$ & $3.492^{-.198}$ & $0.416^{+.046}$ & $0.597^{+.051}$ & $0.713^{+.045}$ & $9.621^{+.647}$ \\
        \bottomrule
    \end{tabular}
    \caption{Comparisons between separate training for each task and joint training for multiple tasks on HumanML3D validation set using MotionGPT-7B. Superscripts indicate the improvement or decrement in the metric. Joint training can achieve better performance for all tasks.}
    \label{tab:separate}
    % \vspace{-0.1cm}
\end{table*}

\begin{table}[t]
    \centering
    \small
    \begin{tabular}{ccc}
        \toprule
             Methods & {Recon $\downarrow$} & {Vel $\downarrow$} \\
        \midrule
            \multicolumn{3}{c}{Initial token} \\
        \midrule
            Text-only & 24.70 & 1.095 \\
            Text + Initial poses & \bf{13.78} & \bf{0.549} \\
        \midrule
            \multicolumn{3}{c}{Last token} \\
        \midrule
            Text-only & 19.70 & 1.172 \\
            Text + Last poses & \bf{6.831} & \bf{0.397} \\
        \midrule
            \multicolumn{3}{c}{Key tokens} \\
        \midrule
            Text-only & 8.035 & 3.813 \\
            Text + Random poses & \bf{5.383} & \bf{2.423} \\
        \bottomrule
    \end{tabular}
    \caption{Evaluation of the effectiveness of pose control conditions on HumanML3D test set using MotionGPT-13B model.}
    \label{tab:pose-consitency}
    % \vspace{-0.2cm}
\end{table}

\subsection{Ablation Study}\label{sec:ablation}
Additionally, extensive ablation studies are conducted on HumanML3D~\cite{guo2022generating} dataset to indicate the effectiveness of our MotionGPT. More ablation studies are included in the supplementary material~\cite{zhang2023motiongpt}.

\paragraph{Capability of pre-trained LLM}
Pre-trained LLMs can provide robust priors about human motion from texts.
In this context, we experiment with base models pre-trained to varying degrees, including LLaMA-7B, LLaMA-13B, and LLaMA without pre-training. 
For the un-pretrained LLaMA, we adopt the same network structure as LLaMA-7B without loading the pre-trained weights. The randomly initialized LLaMA is tuned by LoRA as well, fixing weights during training.
As demonstrated in Tab.~\ref{tab:pretrained-llm}, our results show a strong correlation between the level of pre-training in LLMs and the performance of our model in the text-to-motion generation task. This highlights the significant influence of motion prior extracted from LLM. Note that the training parameters of LoRA are same.

\paragraph{Consistency with pose control conditions}
We demonstrate the effectiveness of pose control conditions by assessing the consistency between pose controls and generated motion on the HumanML3D test set. For each task (initial/last/key), we generate motion with and without pose controls using (text+pose)-to-motion and text-to-motion methods, respectively. The results are shown in Tab.~\ref{tab:pose-consitency}. In comparison to text-only generation, better keyframe pose consistency arises from generating under pose conditions, showcasing (text+pose)-to-motion's effectiveness with pose control.

\paragraph{Comparison with separate training}
To further evaluate the effectiveness of our unified motion generation approach, we conduct separate training for each task on the HumanML3D dataset~\cite{guo2022generating}. The aim is to investigate if multi-task learning could improve the performance of individual control conditions. The comparison results are depicted in Table~\ref{tab:separate}. We find that joint training across all tasks yields significant improvements in all metrics. This effect is especially pronounced when text and last poses are used as conditions. These findings underscore the utility of our unified motion generation approach. It appears that the model's ability to generate motions under a specific control type is boosted by the knowledge derived from other related control conditions.
\section{Conclusion and Limitations}

\paragraph{Conclusion}
This study introduces MotionGPT, a novel method capable of generating human motion using multimodal control signals, such as text and single-frame poses. The approach effectively discretizes pose conditions and creates a unified set of instructions by combining codes from both textual and pose prompts. With MotionGPT, we envision a path toward more practical and versatile motion generation systems, offering a fresh perspective in the field.

\paragraph{Limitations} 
Although current MotionGPT may support any control modalities beyond current human poses and text, this paper only validates the effectiveness on text and human poses.
Validating our MotionGPT on a broader spectrum of possible modalities, such as music pieces, would be highly beneficial to more applications in the real world.

\section{Acknowledgments}
This work is supported by the National Natural Science Foundation of China (Grant No. 62121002 and Grant No. 62272430).

\bibliography{aaai24}
\newpage
% \section*{\scalebox{1.2}{Supplementary Material}}
% In this supplementary material, we provide additional experiments and more visualization results.

\appendix
\setcounter{section}{0}
\renewcommand{\thesection}{\Alph{section}}

\section{Datasets and Evaluation Metrics}
\paragraph{HumanML3D}
HumanML3D~\cite{guo2022generating} is currently the largest 3D human motion-language dataset, paired with well-annotated sequence-level textual descriptions.
It contains 14,616 motion clips and 44,970 descriptions, composed from a vocabulary of 5,371 unique words. The motion sequences, sourced from the AMASS~\cite{mahmood2019amass} and HumanAct12~\cite{guo2020action2motion} datasets, encompass a wide spectrum of human actions, including daily activities, sports, acrobatics, and artistic performances. Each motion clip is accompanied by 3-4 descriptive texts and has been downsampled to 20 fps, with a duration ranging from 2 to 10 seconds. The dataset is partitioned into training, validation, and test sets in an 80\%, 5\%, and 15\% ratio, ensuring no overlap among the subsets.

\paragraph{KIT-ML}
The KIT-ML~\cite{plappert2016kit} dataset is comprised of 3,911 motion sequences along with 6,278 textual descriptions, averaging 9.5 words per description. This dataset is an amalgamation of selected subsets from the KIT WholeBody Human Motion Database~\cite{mandery2015kit} and the CMU Graphics Lab Motion Capture Database~\cite{cmu}. The motion sequences within KIT-ML have been downsampled to a rate of 12.5 fps, ensuring a uniform and manageable rate for analysis and experimentation.

\paragraph{Evaluation metrics}
We introduce new metrics tailored to our proposed motion generation setting, including \textit{Reconstruction Loss (Recon)} and \textit{Velocity Loss (Vel)}, both measured by L2 loss. For scenarios where the initial or final poses are given, the positioning of the corresponding generated poses in the motion sequence is critical. Hence, we propose the use of \textit{Recon} and \textit{Vel} to evaluate the quality of initial or last poses reconstruction and their temporal continuity with neighboring poses. For scenarios where keyframe poses are provided, the positions of the corresponding generated poses within the motion sequence are unknown. Consequently, we calculate the Nearest Euclidean Distance for each key token relative to the corresponding ground truth poses, and report the \textit{Recon} and \textit{Vel} to measure the key poses reconstruction and their temporal continuity with neighboring poses. This approach allows us to quantitatively measure the accuracy of our model in reproducing the provided keyframe poses within the generated motion sequence.

\section{Implementation Details}
\subsection{Motion data pre-processing}
We follow the same data pre-processing method with \cite{guo2022generating}. Specifically, raw 3D motion coordinate is first transformed to make people face the Z+ direction, and subsequently pre-processed into motion features. These features include foot contact, global rotations and translations, local joint positions, velocities, and 6D rotations, having total dimensions of 263 for HumanML3D and 251 for KIT-ML.

\subsection{Training details}
In our experiments, we utilize a frozen 13B LLaMA~\cite{touvron2023llama} model as the foundational LLM, which is subsequently fine-tuned using the LoRA technique. The model training process spans 37,500 epochs, starting with an initial learning rate of 3e-3. We set the batch size to 256, partitioned into micro-batches of 4 to accommodate memory constraints. We employ the AdamW optimizer~\cite{loshchilov2017decoupled} with a weight decay parameter of 0.01 to guide the optimization process. The training duration is approximately 4 hours for the HumanML3D dataset~\cite{guo2022generating} and 3 hours for the KIT-ML dataset~\cite{plappert2016kit} when conducted on a single A100 GPU. These timelines highlight the efficiency of our training process compared to traditional methods. As for the pre-training of motion VQ-VAE~\cite{van2017neural}, we follow the network structure and training strategy of \cite{zhang2023t2m}, which is applied consistently across both datasets.

\section{Additional Experiments}
\label{sec: additional experiments}
To further demonstrate the effectiveness of our model, we conducted several additional experiments on the HumanML3D validation set for text-to-motion generation, employing the MotionGPT-7B model architecture.

\subsection{Hyper-parameters of LoRA}
% During training, all the trainable parameters are sourced from LoRA~\cite{hu2021lora}, which has two hyper-parameters $r$ and $\alpha$. 
During training, all the trainable parameters are sourced from LoRA~\cite{hu2021lora}, which has two hyper-parameters: $r$ and $\alpha$.
% $r$ is the rank of the LoRA parameters, with smaller values indicating fewer parameters. 
% On the other hand, $\alpha$ controls the scale of the outputs from the dense layer of LoRA. The evaluation results are presented in Tab.~\ref{tab:lora}. It is observed that as $r$ increases, the performance of our model improves under almost all metrics while fixing $\alpha$. By maintaining the scale factor $\frac{\alpha}{r}$, akin to the learning rate, it can be demonstrated that increasing $r$ leads to superior performance. When keeping $r$ fixed and adjusting $\alpha$, we discover that setting $\alpha=16$ yields the optimal performance.
% $r$ represents the rank of the LoRA parameters, with smaller values indicating fewer parameters.
% $\alpha$ controls the scale of the outputs from the dense layer of LoRA. 
% The evaluation results are presented in Tab.~\ref{tab:lora}. We observe that, with a fixed $\alpha$, an increase in $r$ leads to improved performance across nearly all metrics. By maintaining the scale factor $\frac{\alpha}{r}$, similar to the learning rate, we demonstrate that a higher $r$ results in superior performance. When keeping $r$ constant and adjusting $\alpha$, we find that setting $\alpha=16$ delivers the best performance.
The rank of LoRA parameters is represented by $r$, with smaller values indicating a fewer number of parameters. 
$\alpha$ controls the scale of the outputs derived from the dense layer of LoRA. As illustrated in Tab.~\ref{tab:lora}, we observe that the performance of our model improves across almost all metrics when we increase the value of $r$, keeping $\alpha$ constant. By maintaining the scale factor $\frac{\alpha}{r}$, which is comparable to the learning rate, we demonstrate that an increase in $r$ leads to superior performance. Additionally, when $\alpha$ is modified while $r$ is kept stable, we find that the optimal performance is achieved when $\alpha$ is set to 16.

\subsection{Evaluation of batch size}
We conducted an evaluation of the performance of the MotionGPT-7B model trained with different batch sizes, and the results are presented in Table~\ref{tab:batch size}. It can be observed that the performances for batch sizes of 128 and 512 are comparable, while the batch size of 256 significantly outperforms the others across nearly all metrics.

\begin{table*}[t]
    \centering
    \begin{tabular}{cc|cccccc}
         \toprule
             \multirow{2}{*}{$r$} & \multirow{2}{*}{$\alpha$} & \multirow{2}{*}{FID $\downarrow$} & \multirow{2}{*}{MM Dist $\downarrow$} & \multicolumn{3}{c}{R-Precision $\uparrow$} & \multirow{2}{*}{Diversity $\uparrow$} \\
             %\cline{2-4}
            & & & & Top-1 & Top-2 & Top-3 & \\
        \midrule
            8 & 16 & 0.837 & 4.142 & 0.315 & 0.491 & 0.600 & 8.847 \\
            16 & 16 & 0.977 & 4.139 & 0.324 & 0.492 & 0.615 & {\bf 9.745} \\
            32 & 16 & \bf{0.576} & 3.982 & 0.330 & 0.507 & 0.618 & 8.801 \\
        \midrule
            8 & 2 & 1.148 & 4.103 & 0.323 & 0.505 & 0.610 & 9.056\\
            16 & 4 & 0.815 & 3.969 & 0.340 & 0.515 & 0.622 & 8.995 \\
            32 & 8 & 0.819 & \underline{3.850} & \underline{0.372} & {\bf 0.555} & \underline{0.652} & \underline{9.420} \\
        \midrule
            64 & 8 & 1.869 & 4.614 & 0.267 & 0.419 & 0.529 & 8.438 \\
            64 & 32 & 0.773 & 4.181 & 0.321 & 0.482 & 0.602 & 8.824 \\
            64 & 16 & \underline{0.590} & \bf{3.796} & \bf{0.376} & \underline{0.553} & \bf{0.657} & 9.048 \\
        \bottomrule
    \end{tabular}
    \caption{Evaluation of text-to-motion generation for {\bf different LoRA parameters} on HumanML3D validation set using MotionGPT-7B. {\bf Bold} and \underline{Underline} indicate the best and the second best result.}
    \label{tab:lora}
\end{table*}

\begin{table*}[ht]
    \centering
    \begin{tabular}{c|cccccc}
         \toprule
             \multirow{2}{*}{Batch Size} & \multirow{2}{*}{FID $\downarrow$} & \multirow{2}{*}{MM Dist $\downarrow$} & \multicolumn{3}{c}{R-Precision $\uparrow$} & \multirow{2}{*}{Diversity $\uparrow$} \\
             %\cline{2-4}
            & & & Top-1 & Top-2 & Top-3 & \\
        \midrule
            128 & 0.752 & 4.063 & 0.314 & 0.491 & 0.612 & {\bf 9.100} \\
            256 & {\bf 0.590} & {\bf 3.796} & {\bf 0.376} & {\bf 0.553} & {\bf 0.657} & 9.048 \\
            512 & 0.684 & 4.010 & 0.311 & 0.495 & 0.611 & 8.947 \\
        \bottomrule
    \end{tabular}
    \caption{Evaluation of text-to-motion generation for MotionGPT-7B training with different batch sizes on HumanML3D validation set.}
    \label{tab:batch size}
\end{table*}

\begin{table*}[h!]
    \centering
    \begin{tabular}{c|cccccc}
         \toprule
             \multirow{2}{*}{Prompts} & \multirow{2}{*}{FID $\downarrow$} & \multirow{2}{*}{MM Dist $\downarrow$} & \multicolumn{3}{c}{R-Precision $\uparrow$} & \multirow{2}{*}{Diversity $\uparrow$} \\
             %\cline{2-4}
            & & & Top-1 & Top-2 & Top-3 & \\
        \midrule
            $V_1$ & 8.506 & 5.490 & 0.200 & 0.331 & 0.447 & 7.566 \\
            $V_2$ & 3.018 & 4.858 & 0.249 & 0.402 & 0.508 & 8.237 \\
            $V_0$ (Ours) & {\bf 0.590} & {\bf 3.796} & {\bf 0.376} & {\bf 0.553} & {\bf 0.657} & {\bf 9.048} \\
        \bottomrule
    \end{tabular}
    \caption{Evaluation of text-to-motion generation for MotionGPT-7B applying different prompts on HumanML3D validation set.}
    \label{tab:prompts}
\end{table*}

\subsection{Evaluation of prompt design}
LLMs are known to be sensitive to prompts, emphasizing the criticality of carefully designing prompts to optimize model performance. In this section, we delve into the impact of employing two alternative prompts and assess their respective performances. Denoting the prompt used in our model as $V_0$, we also introduce two additional prompts, namely $V_1$ and $V_2$, as follows:

\begin{mdframed}[backgroundcolor=gray!20]
\gray{\% Prompts $V_1$} \\
Human motion can be represented by token indices by VQ-VAE. Below is an instruction that describes human motion generation condition types, paired with an input that provides specific conditions. Write a sequence of tokens matching with given conditions.

\textbf{Instruction ($\mathcal{I}):$} \{\gray{Task Prompts:} "Motion description( and the init/last/key pose tokens)."\} \{\gray{Control Conditions:} Text control condition $\mathcal{X}$( <Motion Token> Pose control conditions $\mathcal{P}$ </Motion Token>) \}
\end{mdframed}

\begin{mdframed}[backgroundcolor=gray!20]
\gray{\% Prompts $V_2$} \\
Below is an instruction that describes a task, paired with an input that provides further context. Write a response that appropriately completes the request.

\textbf{Instruction ($\mathcal{I}):$} \{\gray{Task Prompts:} "Generate the token sequence of the given human motion description( under the premise of the given init/last/key pose tokens)."\} \{\gray{Control Conditions:} Text control condition $\mathcal{X}$( <Motion Token> Pose control conditions $\mathcal{P}$ </Motion Token>) \}
\end{mdframed}
For the prompts $V_1$, we incorporated specific human motion generation details into the overall descriptions, while simplifying the task prompts to only include condition types. On the other hand, for the prompts $V_2$, we modified the expression of the task prompts. The comparison results between these prompts are presented in Tab.~\ref{tab:prompts}, highlighting the efficiency and effectiveness of our proposed prompt designs. These findings underscore the significance of well-designed prompts in enhancing the performance of our model.

\section{Qualitative Results}
\label{sec: qualitative results}
\begin{figure*}[ht]
    \centering
    \includegraphics[width=.9\textwidth,trim=230 180 240 60,clip]{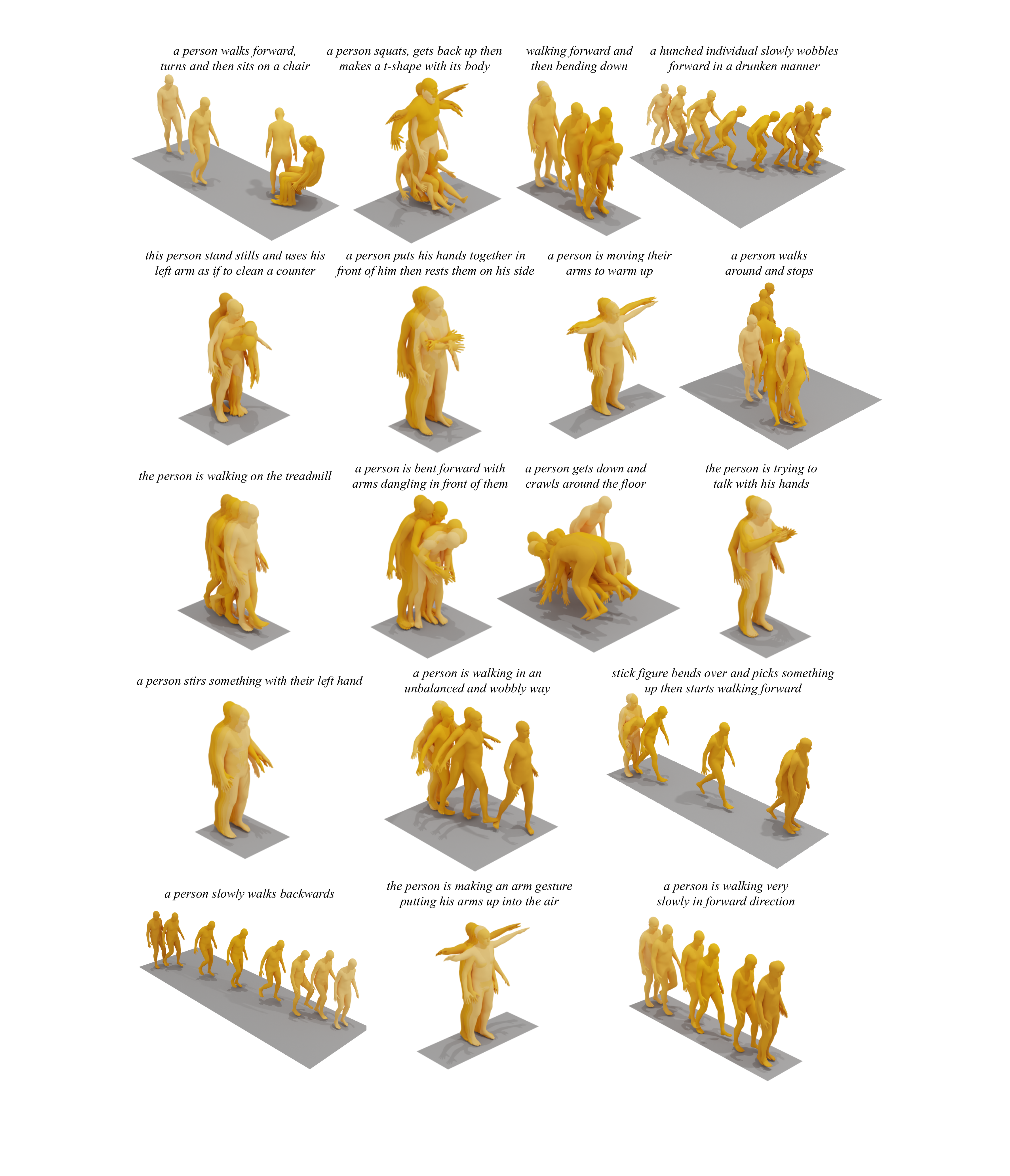}
    \caption{More text-to-motion samples generated by MotionGPT-13B using texts from the HumanML3D test set.}
    \label{fig:t2m}
\end{figure*}

\begin{figure*}[h!]
    \centering
    \includegraphics[width=.85\textwidth,trim=200 200 240 60,clip]{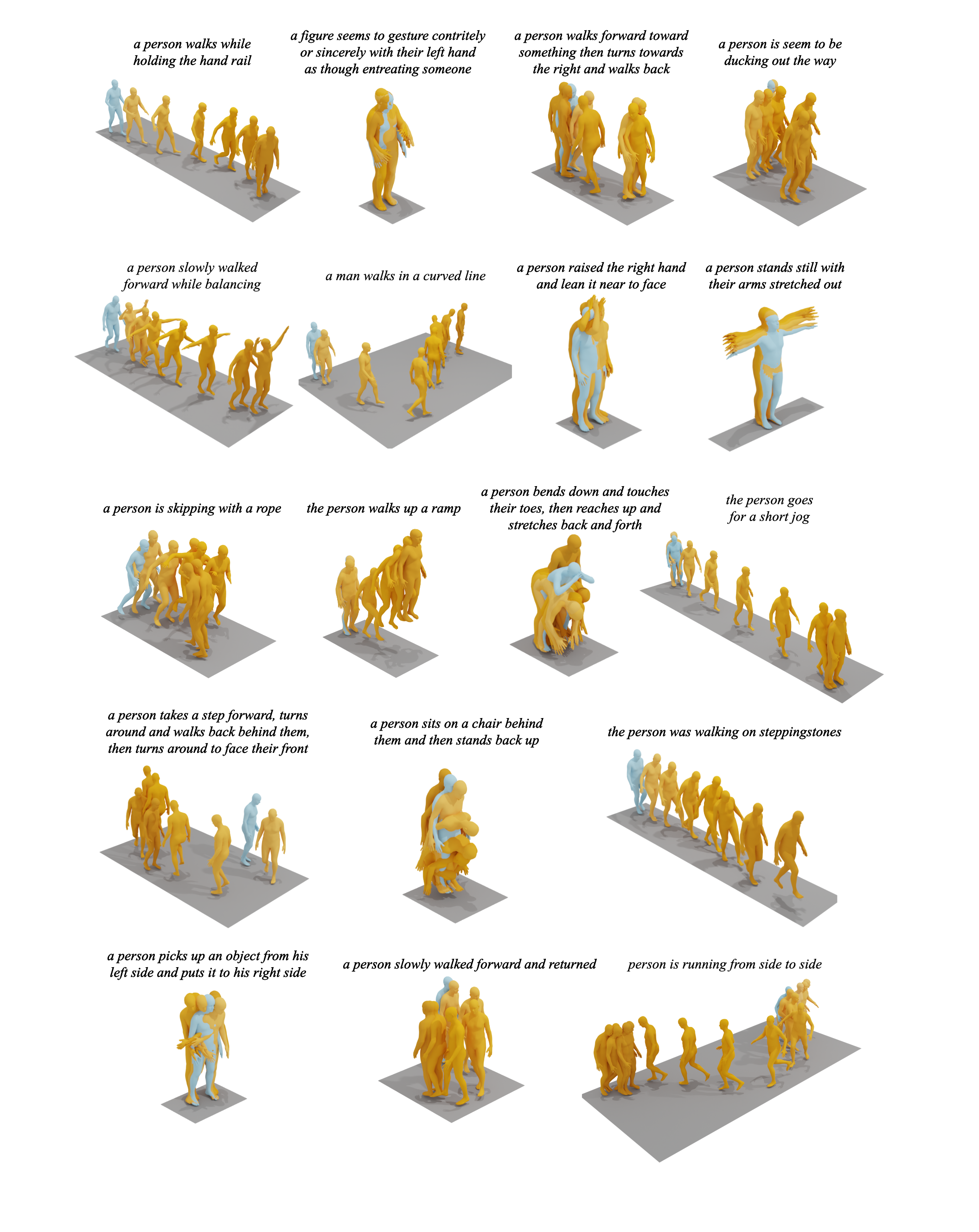}
    \caption{More (text+initial token)-to-motion samples generated by MotionGPT-13B using texts from the HumanML3D test set. The initial pose condition is highlighted in blue.}
    \label{fig:initial}
\end{figure*}

\begin{figure*}[h!]
    \centering
    \includegraphics[width=.8\textwidth,trim=270 220 270 10,clip]{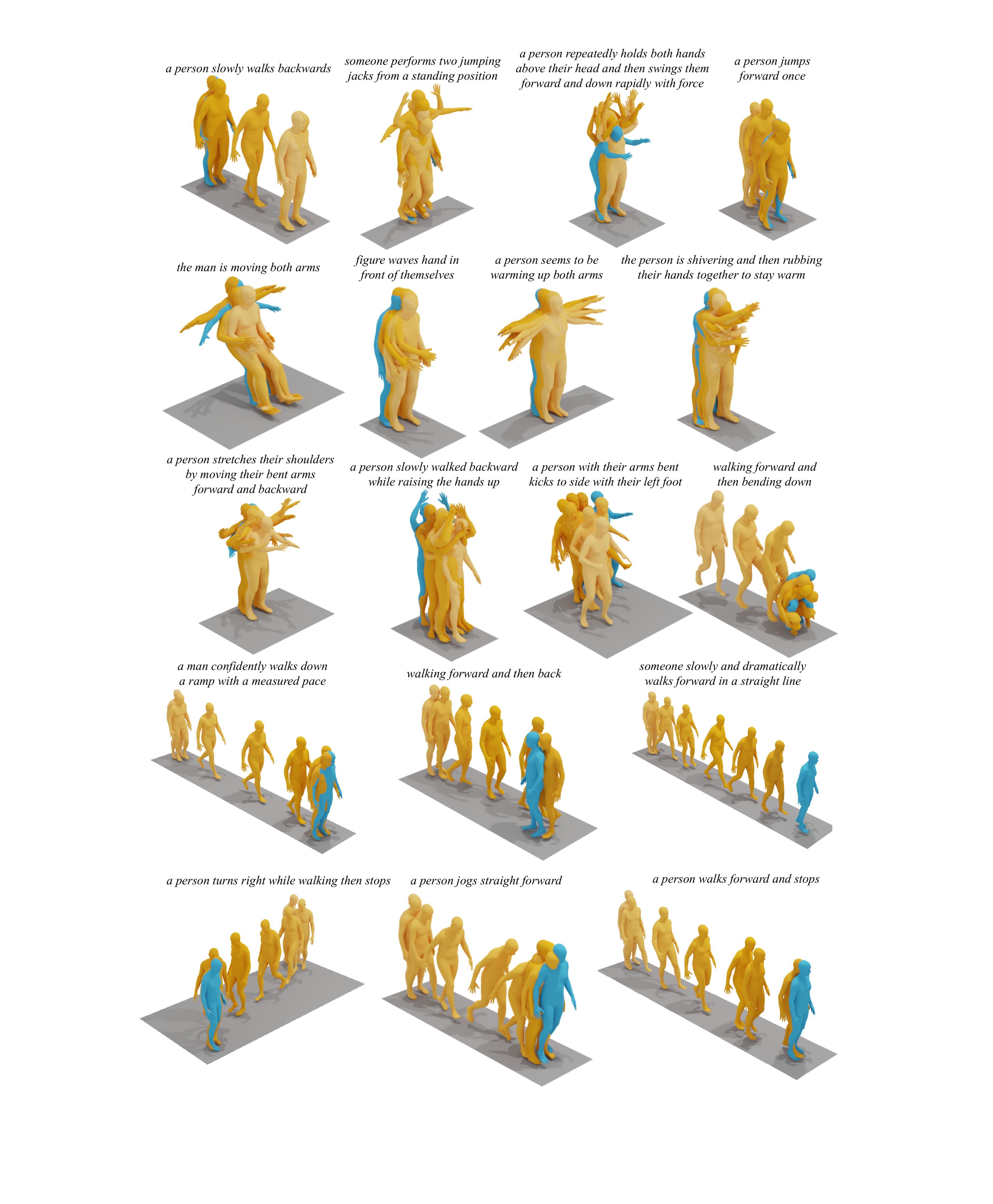}
    \caption{More (text+last token)-to-motion samples generated by MotionGPT-13B using texts from the HumanML3D test set. The last pose condition is highlighted in blue.}
    \label{fig:last}
\end{figure*}

\begin{figure*}[h!]
    \centering
    \includegraphics[width=.8\textwidth,trim=230 150 300 10,clip]{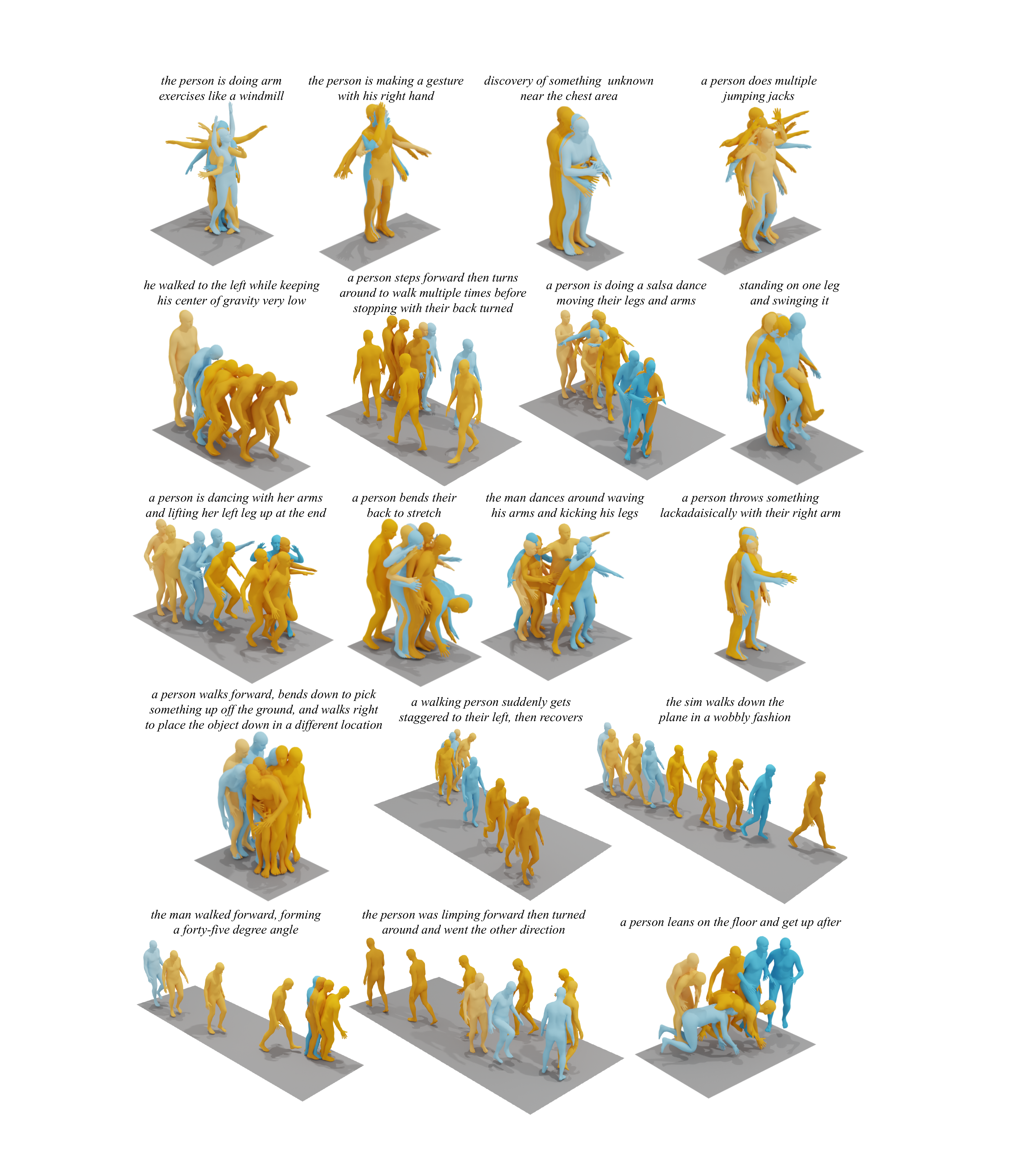}
    \caption{More (text+key tokens)-to-motion samples generated by MotionGPT-13B using texts from the HumanML3D test set. The key pose conditions are highlighted in blue.}
    \label{fig:keys}
\end{figure*}

In this section, we showcase additional qualitative results generated by MotionGPT-13B for all four different control conditions. These results are presented in Figure~\ref{fig:t2m}, Figure~\ref{fig:initial}, Figure~\ref{fig:last}, and Figure~\ref{fig:keys}, respectively. The motion descriptions are sourced from the HumanML3D test set, and the pose control conditions are highlighted in blue. These visual examples offer further insights into the capabilities and performance of our model in generating motions based on different control conditions.

\end{document}